\documentclass[runningheads]{llncs}
\usepackage{booktabs}
\usepackage{multirow}
\usepackage{adjustbox} 

\usepackage{lineno}
\usepackage{xcolor}
\usepackage[T1]{fontenc}
\usepackage{algorithm}
\usepackage{algorithmic}
\usepackage{graphicx,verbatim}
\usepackage{amsmath}
\usepackage{booktabs}

\usepackage{amssymb}
\usepackage{pifont} 
\usepackage{svg}
\usepackage{multirow}  
\usepackage{booktabs}  
\usepackage{graphicx}  
\usepackage{colortbl}

\begin{document}
\title{Robust Mitigation of Age-Dependent Confounding Effects via Sample-Difficulty Decorrelation}
\titlerunning{Confounding Effect Mitigation via Difficulty Decorrelation}
%

\author{
Nikhil Cherian Kurian\inst{1} \and
Victor Caquilpan Parra\inst{1} \and
Abin Shoby\inst{1} \and
Luke Whitbread\inst{1} \and
Lyle J. Palmer\inst{1}
}

\authorrunning{Kurian et al.}

\institute{
Australian Institute for Machine Learning, Adelaide University, Adelaide, Australia\\
\email{\{nikhil.kurian, abin.shoby, luke.whitbread, lyle.palmer\}@adelaide.edu.au}\\
\email{victor.caquilpan@gmail.com}
}
    
\maketitle              
%
\begin{abstract}
Age-dependent performance disparities in medical image classification often arise because age acts as a confounder, linking imaging morphology with disease prevalence. In practice, disparities can manifest as overdiagnosis at ages where disease prevalence is higher and underdiagnosis at ages where prevalence is lower, and can worsen under train–test shifts in the age distribution. Conventional mitigation approaches that enforce strict age invariance may suppress diagnostically meaningful information encoded in age. We therefore propose a robust framework that mitigates the effects of age-dependent confounding by targeting spurious age-linked trends rather than enforcing invariance. Following a warm-up phase, we characterize sample difficulty and model its age-dependent trends in a label-conditioned manner. We decorrelate age from dominant age–difficulty trends using robust, Huber-weighted affinity weights, attenuating confounding-driven shortcuts while preserving clinically meaningful, non-linear age information. We further introduce an Age Coverage Score that scales the decorrelation penalty by mini-batch age variance to ensure stable optimization under limited age diversity. Across two radiology datasets, our approach reduces age-dependent true- and false-positive disparities with minimal AUC impact and remains robust to increasing train–test age distribution shifts.


\end{abstract}
\section{Background and Related Work}
Deep neural networks in medical imaging often rely on shortcuts—readily available cues such as age—to reduce training error rather than learning true disease markers \cite{shortcutlearning2020,xu2024addressing}. Age can act as a confounder because it is associated with both imaging morphology (input appearance) and disease prevalence (label distribution), encouraging models to use age-linked signals instead of subtle pathological features \cite{brown2023detecting,shortcutmedical2024}. The resulting confounding effects appear as systematic age-dependent shifts in operating characteristics, with true- and false-positive rates (TPR/FPR) varying across the age spectrum \cite{glocker2023algorithmic,seyyed2020chexclusion}. Despite this, most age confounding–mitigation approaches discretize age or remove it from the feature space, with limited work examining how such schemes affect model performance~\cite{chu2024strategies}. In this context, enforcing strict age invariance is often undesirable, since age also encodes clinically meaningful variation, motivating approaches that selectively mitigate spurious age dependencies rather than removing age information entirely \cite{gao2025fairread}. These challenges are further exacerbated under train–test shifts in patient age, which commonly accompany prevalence changes \cite{giguere2022fairness}.

We propose a framework that mitigates age-dependent confounding effects by targeting spurious age-linked trends while preserving higher-order age information. Our approach applies a decorrelation that prevents simple age-based trends from distorting learning dynamics, effectively flattening the dominant age-related slope across the age spectrum. This “one-slope” strategy is less sensitive to the particular age distribution in the training set and remains effective under train–test age shifts. To handle limited age diversity within mini-batches, we incorporate an Age Coverage Score that adaptively scales the penalty based on batch-wise age variance. Evaluations on two chest X-ray benchmarks under varying train–test age distribution shifts show that our method reduces age-dependent TPR/FPR disparities while maintaining diagnostic accuracy.

\section{Method}
We consider a radiological finding prediction task and propose a  regularization framework that modulates age-dependent trends in training dynamics.
 Unlike methods that enforce strict age invariance in representations, which may inadvertently discard clinical signal, our approach intervenes on how age influences the learning process. We quantify the training dynamics for each sample $i$ via the gradient magnitude $g_i$. Under standard cross-entropy, $g_i$ (with respect to the logit) is equivalent to the per-sample absolute prediction error $|p_i - y_i|$, where $p_i\in[0,1]$ is the predicted probability and $y_i\in\{0,1\}$ is the ground-truth label~\cite{murphy2012machine}. This provides a bounded $[0,1]$ proxy for sample difficulty that remains computationally efficient, requiring no gradient hooks or explicit backward-pass interventions.

\begin{figure}[t]
    \centering
    \includegraphics[width=\textwidth]{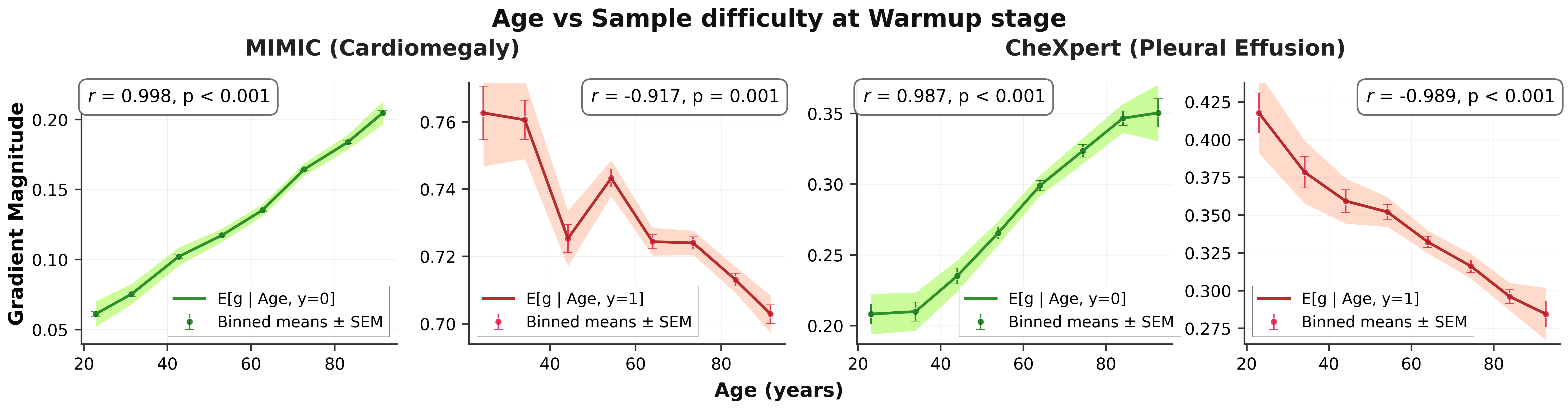}
\caption{\textbf{Empirical age--difficulty trends at warm-up.}
Age-conditioned mean sample difficulty $g$ at warm-up, averaged over five seeds.
Points show age-bin means ($\pm$SEM); curves show $\mathbb{E}[g \mid \text{Age}, y]$ (red: $y{=}1$, green: $y{=}0$), matching the observed lower TPR/FPR at younger ages and higher TPR/FPR at older ages~\cite{glocker2023algorithmic,seyyed2020chexclusion}.}

   \label{fig1:main_results}
\end{figure}

\subsection{Empirical Motivation}
Following a model warm-up phase (one-epoch), we analyze the conditional expectation $\mathbb{E}[g \mid \mathrm{Age}, y]$ and observe a strong linear dependence on age across the radiology datasets (Fig.~\ref{fig1:main_results}), with Pearson correlation coefficients $|r| > 0.9$. Specifically, when disease is present ($y=1$), samples from older patients are, on average, "easier" (lower $g_i$) for the model than those from younger patients; conversely, when disease is absent ($y=0$), older patients represent more difficult samples. These trends mirror the empirical age-dependent confounding effects in these datasets—namely, reduced true-positive rates (under-diagnostic) for younger patients and elevated false-positive rates (over-diagnostic) for older patients—suggesting the model over-relies on age as a predictive shortcut~\cite{glocker2023algorithmic}.

Although the strength and direction of this trend may vary across cohorts (e.g., pediatric vs.\ geriatric dominant), the consistent presence of monotonic age--difficulty relationships can distort training dynamics and induce systematic bias~\cite{glocker2023algorithmic}. This motivates an intervention that suppresses spurious age-based shortcuts responsible for these disparities, rather than enforcing strict age invariance that may remove clinically meaningful information~\cite{xu2024addressing}.

\subsection{Weighted Slope Penalty}
Motivated by the observed label-conditioned age--difficulty relationship, we model the interaction between normalized age $z_i$ and sample difficulty $g_i$ as:
\begin{equation}
g_i \;\approx\; \alpha_{y_i} + \beta_{y_i} z_i ,
\label{eq:age_trend}
\end{equation}
where parameters are estimated independently for each label $y_i \in \{0,1\}$. Rather than enforcing full statistical independence, we impose a weaker constraint by decorrelating $z$ and $g$. This is implemented by penalizing the squared estimate of the age--difficulty slope $\beta_{y_i}$ within each mini-batch. This approach suppresses dominant linear age-based shortcuts without enforcing full age invariance.

Directly penalizing $\beta_{y_i}^{2}$ (decorrelation) is often unstable, as $g_i$ inherently reflects clinical variability and annotation noise. Therefore, unweighted decorrelation risks penalizing highly informative samples or inducing optimization instability. 
We address this by estimating slopes via a Weighted Ordinary Least Squares (WOLS) formulation, where each sample is weighted by its \emph{trend affinity}---a measure of alignment with the dominant age--difficulty relationship.
Trend affinity is estimated once immediately after warm-up using label-conditioned Huber regression on the full training set and then kept fixed, since early gradients best reflect relative difficulty while later gradients shrink and yield noisier re-estimates~\cite{arpit2017closer}. Moreover, Huber regression residuals at convergence provide per-sample weights equivalent to a WOLS solution \cite{huber1964robust,hampel1986robust}. Specifically, for each sample $i$, we compute the residual $r_i$ from the Huber fit:
\begin{equation}
r_i \;=\; g_i - \bigl(\hat{\alpha}_{y_i} + \hat{\beta}_{y_i} z_i \bigr),
\end{equation}
which measures the deviation from the fitted trend. Influence weights $w_i$ are then derived from a robust scale estimate $\delta_{y_i}$ of the residuals:
\begin{equation}
w_i \;=\;
\begin{cases}
1, & |r_i| \le \delta_{y_i}, \\[4pt]
\dfrac{\delta_{y_i}}{|r_i|}, & |r_i| > \delta_{y_i}.
\end{cases}
\end{equation}


Samples within a central residual band receive full weight, while those farther from the trend are smoothly downweighted. Consequently, mini-batch slope penalization primarily acts on samples participating in the systematic linear age--difficulty dependence, reducing the risk of unstable updates or training collapse when decorrelating across all gradients. In our experiments, $\delta_{y_i}$ was set to the median absolute deviation (MAD) of the residuals, a standard robust scale estimator in Huber-type regression and weighted least squares~\cite{huber1964robust,hampel1986robust}.

\subsection{Batch-level Slope Estimation}
During optimization, we estimate the linear age--difficulty trend within each mini-batch $\mathcal{B}$, calculated separately for each class $y \in \{0,1\}$ using the trend-affinity weights $w_i$. Let $\mathcal{B}_y = \{ i \in \mathcal{B} \mid y_i = y \}$ denote the subset of batch samples with label $y$. The WOLS estimate of the batch-level slope, $\hat{\beta}_{y}$, is given by the ratio of the weighted age–difficulty covariance to the weighted age variance~\cite{hastie2009elements}:

\begin{equation}
\hat{\beta}_{y}
\;=\;
\frac{\sum_{i \in \mathcal{B}_y} w_i \,(z_i - \mu_{z,y})\,(g_i - \mu_{g,y})}
     {\sum_{i \in \mathcal{B}_y} w_i \,(z_i - \mu_{z,y})^2},
\end{equation}

where $\mu_{z,y}$ and $\mu_{g,y}$ are the weighted batch means of normalized age and sample difficulty:

\begin{equation}
\mu_{z,y}
\;=\;
\frac{\sum_{i \in \mathcal{B}_y} w_i z_i}{\sum_{i \in \mathcal{B}_y} w_i},
\qquad
\mu_{g,y}
\;=\;
\frac{\sum_{i \in \mathcal{B}_y} w_i g_i}{\sum_{i \in \mathcal{B}_y} w_i}.
\end{equation}
We define the corresponding trend regularization term for each class as:
\begin{equation}
\mathcal{L}_{\text{slope},y} \;=\; \hat{\beta}_{y}^{2}.
\end{equation}

By minimizing this penalty, the model is encouraged to reduce reliance on dominant linear age-based shortcuts linked to age during training. Because $g_i$ is derived from the model output $p_i$, the penalty is fully differentiable, allowing for direct modulation of training dynamics.

\subsection{Age Coverage Modulation}
To account for the reliability of batch-wise slope estimation, we introduce an \emph{Age Coverage Score} $C \in [0,1]$, which quantifies the age diversity within a mini-batch. We define $C$ as the normalized variance of the normalized age values $\{z_i\}_{i=1}^N$:
\begin{equation}
\mathrm{Var}(z) \;=\; \frac{1}{N} \sum_{i=1}^{N} (z_i - \bar{z})^2,
\qquad
C \;=\; \frac{\mathrm{Var}(z)}{\mathrm{Var}_{\max}},
\end{equation}
where $N$ is the mini-batch size, $\bar{z}$ is the batch mean and $\mathrm{Var}_{\max}=0.25$ is the theoretical maximum variance for $z\in[0,1]$~\cite{bhatia2000better}. Higher $C$ indicates broader age coverage and more reliable slope estimation within a mini-batch.

We incorporate this score into a coverage-aware objective:
\begin{equation}
\mathcal{L}_{\text{total}}
\;=\;
\mathcal{L}_{\text{BCE}}
\;+\;
\lambda \sum_{y \in \{0,1\}} C_y \,\mathcal{L}_{\text{slope},y}
\end{equation}
where $C_y$ is the coverage score computed over batch samples with label $y$. This modulation downweights decorrelation updates when age coverage is limited, improving robustness under age imbalance and train--test age distribution shifts. Consequently, this coverage-aware modulation ensures that the model only reduces the age--difficulty slope toward zero when the mini-batch provides a representative sample of the underlying age distribution. By basing the penalty on sufficient evidence of a trend, the framework maintains stable fairness mitigation that is inherently less sensitive to the specific demographic compositions encountered during train--test distribution shifts.

\section{Experiment Details}
\begin{figure}[t]
    \centering
     \includegraphics[scale=0.08]
{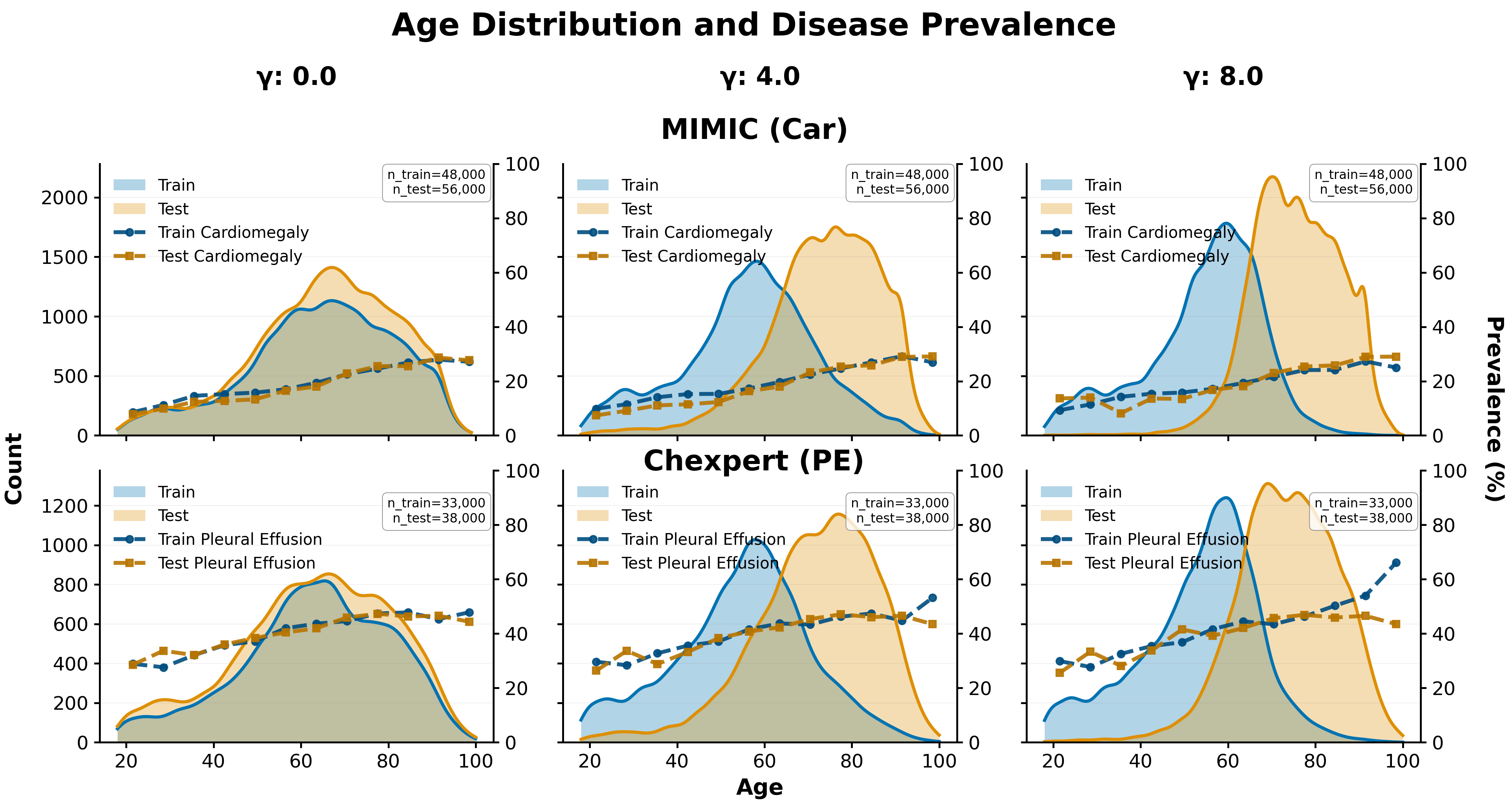}
     \caption{\textbf{Sampling-induced age distribution shifts.} Training (blue) and test (orange) age densities for $\gamma \in \{0,4,8\}$. Larger $\gamma$ yields stronger age shifts. Dashed lines show prevalence as a function of age in train (blue) and test (orange).}

    \label{fig:distribution_shift}
\end{figure}

\subsection{Datasets and Distribution Shift}
We evaluate our framework on the most prevalent findings in two benchmark radiology datasets: \textbf{CheXpert Pleural Effusion (CXP PE)}~\cite{irvin2019chexpert} and \textbf{MIMIC Cardiomegaly (MIMIC Car)}~\cite{MIMIC-CXR2019}, which exhibit some of the strongest age-related unfairness~\cite{glocker2023algorithmic}. To conduct experiments with shifted age distributions between training and testing environments, we first establish a strict patient-level split. We then apply an age-based exponential sampling scheme to generate refined splits with skewed age profiles. For a sample with normalized age $z \in [0, 1]$, the sampling weights for the training and testing pools ($w_{tr}$ and $w_{te}$) are:
\begin{equation}
w_{tr} \propto e^{\gamma(z - b_{tr})}, \quad w_{te} \propto e^{-\gamma(z - b_{te})},
\end{equation}
where the training pivot $b_{tr}$ is set at the 25th percentile and the testing pivot $b_{te}$ at the 75th percentile of the dataset age distribution. 

As shown in Fig.~\ref{fig:distribution_shift}, we utilize three shift intensities $\gamma \in \{0, 4, 8\}$. As $\gamma$ increases, the training set is drawn increasingly from a younger population while the test set shifts toward an older cohort. This creates a challenging stress test scenario with significant covariate shift, as the model is evaluated on a population with higher disease prevalence and increased morphological complexity than observed during training. Finally, 10\% of the training pool is reserved as a distribution-consistent validation set for hyperparameter tuning.

\subsection{Experimental Setting}

We use DenseNet-121~\cite{huang2017densely} for all experiments across both datasets. Models are trained for 30 epochs on a single NVIDIA GeForce RTX 3090 with batch size 64 and intial learning rate 0.001 using Adam optimizer~\cite{kingma2014adam}. Results are aggregated over five random seeds. The test operating point is selected by maximizing Youden’s J on the validation set~\cite{glocker2023algorithmic,gao2025fairread}. The regularization coefficient $\lambda$ is set to 1.2, selected using validation
data to balance fairness metrics against overall AUC~\cite{menon2018cost,gao2025fairread}.

\subsection{Evaluation Metrics}

To evaluate mitigation, we assess fairness improvement using a continuous-age analogue of Equalized Odds~\cite{zafar2017fairness} that captures age-dependent TPR and FPR disparities at a fixed operating point selected on the validation set. Following prior work~\cite{brown2023detecting,yang2024limits}, we use the \emph{separation} metric to quantify systematic age-related performance disparities.

Age dependence is estimated by fitting logistic models of age separately to
positive ($y=1$) and negative ($y=0$) samples, yielding coefficients $s^{+}$
and $s^{-}$ that capture structured age-dependent variation in TPR and FPR,
respectively. We report $|s^{+}|$ and $|s^{-}|$ as the magnitudes of these effects. Further, the separation coefficient is
defined as $\text{Sep}=\tfrac{1}{2}(|s^{+}|+|s^{-}|)$. For interpretability, we
report fairness as the relative change per decade of age~\cite{brown2023detecting},
\begin{equation}
\Delta \mathrm{Sep}_{10} = \exp(\text{Sep}\cdot\Delta a) - 1, \qquad \Delta a=10~\text{years}.
\end{equation}

We further analyze the AUC (and $\Delta$AUC)–$\Delta \mathrm{Sep}_{10}$ trade-off in the classical
fairness–accuracy setting~\cite{menon2018cost}, particularly as $\gamma$ increases.

\subsection{Comparison with Existing Methods}
Our approach is compared against Empirical Risk Minimization (ERM) and other  baselines: JTT \cite{JTT2021}, which upweights samples misclassified in an initial training stage; Resampled Batch \cite{brown2023detecting}, which enforces a uniform age distribution per mini-batch; 
GroupDRO~\cite{DRO2020}, which
performs worst-group reweighting over age-defined groups (10-year age bins,
matching the $\Delta \mathrm{Sep}_{10}$ metric)
and Adversarial Training (Adv) \cite{brown2023detecting}, uses a gradient-reversal adversarial head trained with a mean squared error objective.
\section{Results and Discussion}
\begin{table}[t]
\centering
\caption{Age-separation metrics ($|S^{+}|, |S^{-}|, \Delta \text{Sep}_{10}$) across shift strengths $\gamma \in \{0, 4, 8\}$. Lower values indicate superior fairness. $|S^{\pm}|$ values are scaled by $10^{-2}$ and averaged over five runs with standard error. \textbf{Bold} denotes the best result; \underline{underlined} the second best.}
\label{tab:sep10_across_gamma}
\small
\setlength{\tabcolsep}{3.5pt}
\renewcommand{\arraystretch}{1.15}
\newcommand{\sd}[1]{{\scriptsize $\pm$#1}}
\begin{adjustbox}{max width=\linewidth}
\begin{tabular}{llcccccccccc}
\toprule
\multirow{2}{*}{Method} & \multirow{2}{*}{Dataset} &
\multicolumn{3}{c}{$\gamma=0$} &
\multicolumn{3}{c}{$\gamma=4$} &
\multicolumn{3}{c}{$\gamma=8$} &
\multirow{2}{*}{\shortstack{Avg \\ $\Delta \mathrm{Sep}_{10}$ (\%)}} \\
\cmidrule(lr){3-5}\cmidrule(lr){6-8}\cmidrule(lr){9-11}
 &  & $|S^{+}|$ & $|S^{-}|$ & $\Delta \mathrm{Sep}_{10} (\%)$ & $|S^{+}|$ & $|S^{-}|$ & $\Delta \mathrm{Sep}_{10} (\%)$ & $|S^{+}|$ & $|S^{-}|$ & $\Delta \mathrm{Sep}_{10} (\%)$ &  \\
\midrule
ERM       & \multirow{6}{*}{\centering\shortstack{MIMIC\\(Car)}}
  & 2.31\sd{0.04} & 3.58\sd{0.03} & 32.21\sd{0.27}
  & 2.11\sd{0.05} & 3.49\sd{0.06} & 32.26\sd{0.20}
  & 2.04\sd{0.02} & 3.54\sd{0.06} & 32.31\sd{0.21} & 32.26\sd{0.22} \\
ReSampled &
  & 2.00\sd{0.05} & 3.40\sd{0.03} & 30.90\sd{0.36}
  & 2.42\sd{0.08} & 3.75\sd{0.07} & 36.13\sd{0.31}
  & 2.69\sd{0.04} & 3.75\sd{0.04} & 38.04\sd{0.35} & 35.02\sd{0.32} \\
Adv       &
  & 1.31\sd{0.02} & \underline{2.62\sd{0.06}} & 21.73\sd{0.32}
  & 1.95\sd{0.05} & 3.44\sd{0.03} & 30.95\sd{0.30}
  & 2.04\sd{0.06} & 3.54\sd{0.04} & 32.22\sd{0.26} & 28.30\sd{0.26} \\
JTT       &
  & 1.42\sd{0.03} & 2.92\sd{0.05} & 24.23\sd{0.25}
  & \underline{1.67\sd{0.03}} & \underline{3.31\sd{0.07}} & \underline{28.27\sd{0.31}}
  & \underline{1.97\sd{0.02}} & \underline{3.32\sd{0.06}} & \underline{30.28\sd{0.28}} & \underline{27.60\sd{0.28}} \\
GDRO      &
  & \underline{1.01\sd{0.03}} & 2.84\sd{0.04} & \underline{21.26\sd{0.30}}
  & 1.67\sd{0.04} & 3.36\sd{0.03} & 28.59\sd{0.27}
  & 2.42\sd{0.03} & 3.72\sd{0.04} & 35.93\sd{0.34} & 28.59\sd{0.27} \\
Our       &
  & \textbf{0.84\sd{0.03}} & \textbf{2.42\sd{0.05}} & \textbf{17.69\sd{0.31}}
  & \textbf{1.02\sd{0.03}} & \textbf{2.84\sd{0.06}} & \textbf{21.30\sd{0.28}}
  & \textbf{1.36\sd{0.05}} & \textbf{2.86\sd{0.06}} & \textbf{23.48\sd{0.25}} & \textbf{20.82\sd{0.24}} \\
\midrule
ERM       & \multirow{6}{*}{\centering\shortstack{CXP\\(PE)}}
  & 1.35\sd{0.02} & 1.89\sd{0.03} & 17.59\sd{0.14}
  & 1.30\sd{0.03} & 1.99\sd{0.06} & 17.89\sd{0.14}
  & 1.59\sd{0.02} & 1.76\sd{0.05} & 18.21\sd{0.19} & 17.90\sd{0.17} \\
ReSampled &
  & 1.00\sd{0.02} & \underline{1.33\sd{0.03}} & 12.30\sd{0.17}
  & \underline{0.92\sd{0.03}} & 1.47\sd{0.05} & \underline{12.69\sd{0.16}}
  & 1.30\sd{0.01} & 1.53\sd{0.02} & 15.58\sd{0.21} & 13.52\sd{0.21} \\
Adv       &
  & 0.86\sd{0.03} & 1.37\sd{0.06} & \underline{11.82\sd{0.15}}
  & 1.06\sd{0.03} & \underline{1.39\sd{0.07}} & 13.01\sd{0.18}
  & 1.41\sd{0.02} & 1.84\sd{0.06} & 17.60\sd{0.23} & 14.14\sd{0.19} \\
JTT       &
  &\underline{0.73\sd{0.03}} & 1.50\sd{0.05} & 11.82\sd{0.24}
  & 1.10\sd{0.04} & 1.40\sd{0.04} & 13.28\sd{0.21}
  & \textbf{0.81\sd{0.02}} & 1.69\sd{0.07} & \underline{13.29\sd{0.24}} & \underline{12.80\sd{0.21}} \\
GDRO      &
  & 0.94\sd{0.01} & 1.41\sd{0.05} & 12.46\sd{0.16}
  & 1.12\sd{0.03} & 1.59\sd{0.05} & 14.55\sd{0.19}
  & 1.52\sd{0.04} & \underline{1.52\sd{0.06}} & 16.44\sd{0.22} & 14.49\sd{0.22} \\
Our       &
  & \textbf{0.56\sd{0.02}} & \textbf{0.89\sd{0.07}} &  \textbf{7.51\sd{0.14}}
  & \textbf{0.54\sd{0.03}} & \textbf{0.85\sd{0.06}} &  \textbf{7.19\sd{0.15}}
  & \underline{0.88\sd{0.03}} & \textbf{0.80\sd{0.06}} &  \textbf{8.76\sd{0.23}} &  \textbf{7.82\sd{0.18}} \\
\bottomrule
\end{tabular}
\end{adjustbox}
\end{table}

Table~\ref{tab:sep10_across_gamma} reports $\Delta \mathrm{Sep}_{10} (\%)$, defined as the percentage relative change in TPR and FPR per decade of age, across shift strengths $\gamma \in {0,4,8}$ (lower is better).
Across both datasets, our method achieves the lowest $\Delta \mathrm{Sep}_{10}$ at all $\gamma$, indicating consistently reduced age dependence.

On MIMIC (Car) and CXP (PE), our method attains average $\Delta \mathrm{Sep}_{10} (\%)$ of $20.82\%$ and $7.82$\%, yielding relative reductions of $\sim$35\% and $\sim$56\% over ERM ($32.26\%$, $17.90\%$).
With increasing age shift, several mitigation baselines degrade substantially; notably, age-invariant adversarial training on MIMIC worsens from $21.73\%$ ($\gamma{=}0$) to $32.22\%$ ($\gamma{=}8$), approaching ERM-level fairness and in some cases matching or underperforming ERM.
In contrast, our method degrades more gradually ($17.69\% \to 23.48\%$ on MIMIC) and remains comparatively stable on CXP ($7.51\% \to 8.76\%$). 

\begin{figure}[t]
    \centering
    \includegraphics[scale=0.25]
{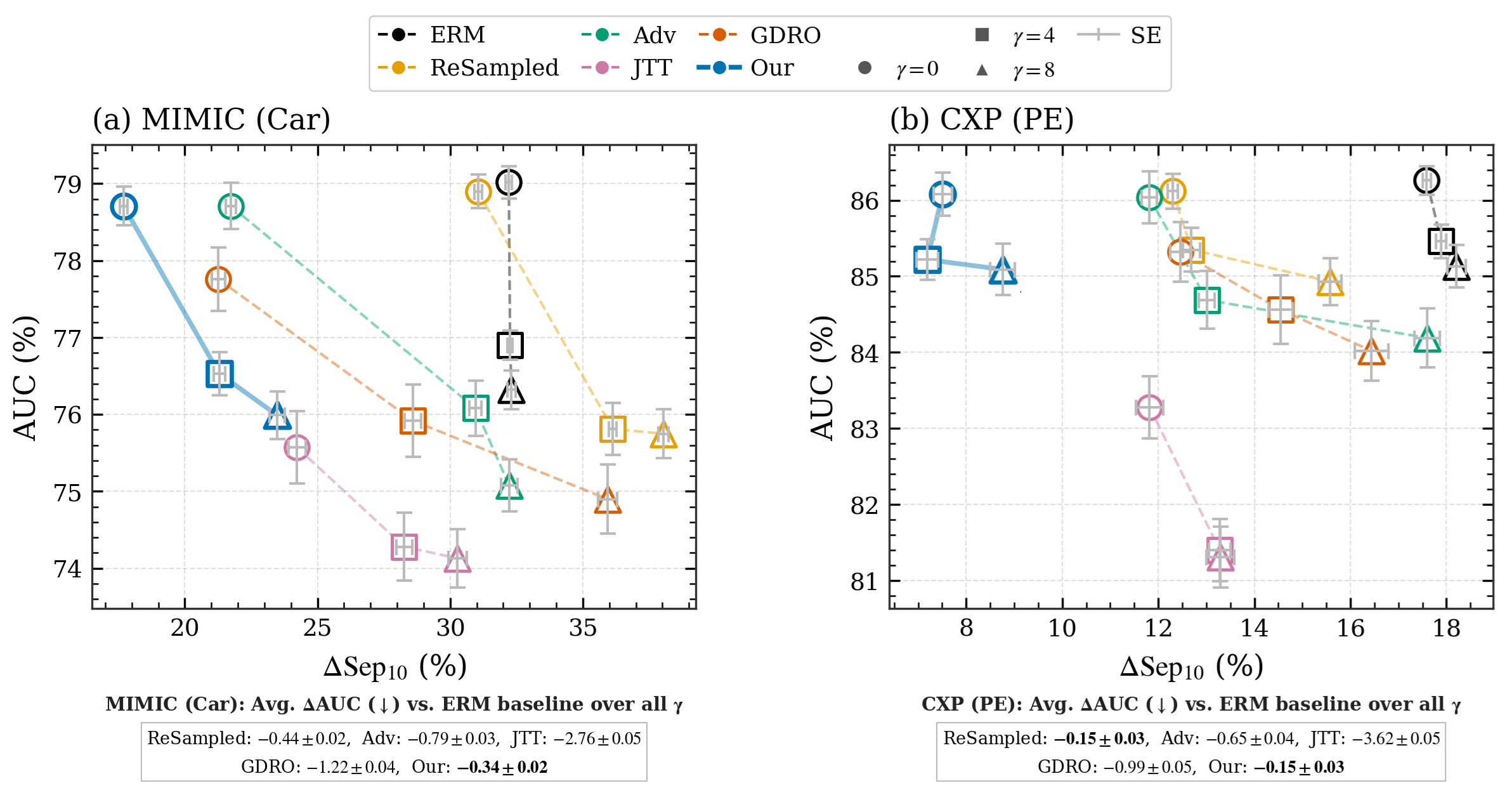}
\caption{AUC (\%) vs.\ $\Delta\mathrm{Sep}_{10}(\%)$ on two datasets. Marker shape encodes $\gamma\in\{0,4,8\}$; error bars show Standard Error\@. Our method (blue) achieves the best fairness--performance trade-off. $\Delta$AUC relative to ERM AUC (mean\,$\pm$\,SE, avg.\ over $\gamma$) below.}
    \label{fig:auc_sep10}
\end{figure}

Fig.~\ref{fig:auc_sep10} shows the AUC--$\Delta \mathrm{Sep}_{10}$ trade-off.
All methods exhibit expected AUC decline with increasing $\gamma$, but most fairness baselines incur large performance losses (e.g., JTT: $-2.76 \pm 0.05$ on MIMIC; $-3.62 \pm 0.05$ on CXP).
Our method achieves the most favorable trade-off, consistently attaining the lowest $\Delta \mathrm{Sep}_{10}$ with the smallest AUC drop ($-0.34 \pm 0.02$ on MIMIC; $-0.15 \pm 0.03$ on CXP). Table~\ref{tab:ablation} evaluates trend affinity and coverage on MIMIC.
Removing either component degrades fairness and slightly reduces AUC, confirming complementarity.
Eliminating coverage causes the largest fairness drop (Avg.~$\Delta\text{Sep}_{10}(\%)=22.80\%$), while removing trend affinity yields a smaller but consistent degradation ($21.25\%$), both increasingly pronounced under stronger shift.
Neither ablation matches the full model.

\begin{table}[t]
\centering
\caption{Ablation study evaluated with $\Delta \mathrm{Sep}_{10} (\%)$ and AUC (\%) across shift strengths $\gamma \in \{0,4,8\}$, examining the contributions of trend affinity and coverage score. The Reference row corresponds to the full model with both components enabled.}
\label{tab:ablation}
\small
\setlength{\tabcolsep}{3.5pt}
\renewcommand{\arraystretch}{1.15}
\newcommand{\sd}[1]{{\scriptsize $\pm$#1}}
\begin{adjustbox}{max width=\linewidth}
\begin{tabular}{lccccccccc}
\toprule
\multirow{2}{*}{Method} &
\multicolumn{2}{c}{$\gamma=0$} &
\multicolumn{2}{c}{$\gamma=4$} &
\multicolumn{2}{c}{$\gamma=8$} &
\multirow{2}{*}{\shortstack{Avg \\ $\Delta \mathrm{Sep}_{10}$ (\%)}} &
\multirow{2}{*}{\shortstack{Avg \\ AUC (\%)}} \\
\cmidrule(lr){2-3}\cmidrule(lr){4-5}\cmidrule(lr){6-7}
 & $\Delta \mathrm{Sep}_{10} (\%)$ & AUC (\%) &
   $\Delta \mathrm{Sep}_{10}$ (\%)& AUC (\%) &
   $\Delta \mathrm{Sep}_{10}$ (\%)& AUC (\%) & & \\
\midrule
W/O Trend Affinity
  & 18.02\sd{0.23} & 78.24\sd{0.25}
  & 21.71\sd{0.28} & 75.98\sd{0.22}
  & 24.01\sd{0.29} & 75.18\sd{0.26}
  & 21.25\sd{0.32} & 76.47\sd{0.24} \\
W/O Coverage Score
  & 19.41\sd{0.26} & 78.81\sd{0.21}
  & 23.87\sd{0.31} & 76.31\sd{0.18}
  & 25.13\sd{0.27} & 75.53\sd{0.29}
  & 22.80\sd{0.29} & 76.88\sd{0.23} \\
Reference
  & \textbf{17.69\sd{0.23}} & \textbf{78.99\sd{0.21}}
  & \textbf{21.30\sd{0.26}} & \textbf{76.53\sd{0.25}}
  & \textbf{23.48\sd{0.28}} & \textbf{75.98\sd{0.27}}
  & \textbf{20.82\sd{0.24}} & \textbf{77.17\sd{0.24}} \\
\bottomrule
\end{tabular}
\end{adjustbox}
\end{table}

\section{Conclusion}
Age-dependent confounding is a critical source of systematic bias in medical image classification, where models exploit age-linked shortcuts that produce clinically harmful over- and under-diagnostic disparities and worsen under distribution shift.We address this by attenuating linear age–difficulty shortcuts without enforcing strict age invariance. Across two radiology datasets, our framework reduces age-dependent TPR/FPR disparities with minimal AUC impact and provides robust fairness mitigation under increasing train–test age distribution shifts.
\section*{Acknowledgements}
This work was supported by GSK AI through the Responsible AI Research Grant. The authors also acknowledge the Australian Institute for Machine Learning for providing institutional support and research infrastructure.

\bibliography{references}
\bibliographystyle{splncs04}

%
%
%

\end{document}